%% file: sparse_motion.tex
\documentclass[10pt,twocolumn,letterpaper]{article}
\pdfoutput=1
\usepackage{cvpr}
\usepackage{times}
\usepackage{epsfig}
\usepackage{graphicx}
\usepackage{amsmath}
\usepackage{amssymb}
\usepackage{multirow}
\usepackage{commath}
\usepackage{pdfpages}

\usepackage[pagebackref=true,breaklinks=true,letterpaper=true,colorlinks,bookmarks=false]{hyperref}

\DeclareMathOperator*{\argmin}{\arg\!\min}

\cvprfinalcopy 

\begin{document}

\title{Sparse Representations for Object and Ego-motion Estimation in Dynamic Scenes}

\author{Hirak J Kashyap$^1$, Charless Fowlkes$^1$, Jeffrey L Krichmar$^{1,2}$\\
$^1$Department of Computer Science, $^2$Department of Cognitive Sciences\\
University of California, Irvine\\
{\tt\small kashyaph@uci.edu, fowlkes@ics.uci.edu, jkrichma@uci.edu}
}

\maketitle

\begin{abstract}
Dynamic scenes that contain both object motion and egomotion are a challenge for monocular visual odometry (VO).
Another issue with monocular VO is the scale ambiguity, i.e. these methods cannot estimate scene depth and camera motion in real scale. Here, we propose a learning based approach to predict camera motion parameters directly from optic flow, by marginalizing depthmap variations and outliers. This is achieved by learning a sparse overcomplete basis set of egomotion in an autoencoder network, which is able to eliminate irrelevant components of optic flow for the task of camera parameter or motionfield estimation.
The model is trained using a sparsity regularizer and a supervised egomotion loss, and achieves the state-of-the-art performances on trajectory prediction and camera rotation prediction tasks on KITTI and Virtual KITTI datasets, respectively. The sparse latent space egomotion representation learned by the model is robust and requires only 5\% of the hidden layer neurons to maintain the best trajectory prediction accuracy on KITTI dataset. Additionally, in presence of depth information, the proposed method demonstrates faithful object velocity prediction for wide range of object sizes and speeds by global compensation of predicted egomotion and a divisive normalization procedure.
\end{abstract}

\section{Introduction}
\input{section_1.tex}



\section{Related work}
\input{section_2.tex}

\input{section_3.tex}

\section{The motion field generator network}
\input{section_4.tex}
\section{Experiments}
\input{section_5.tex}

\section{Conclusion}
\input{section_6.tex}

{\small
\bibliographystyle{ieee}
\bibliography{sparse_motion}
}



\section*{Supplementary material}

# Supplementary Material

In this supplementary file, we first provide proof of elimination of irrelevant variances in the input optic flow by MFG for prediction of egomotion. Also, additional qualitative results for object motion perception and egomotion representation by the hidden layer neurons are presented.

## 1. Elimination of irrelevant information by MFG

One of the key operations performed by the proposed MFG network for egomotion prediction is the elimination of irrelevant information, such as depth variations, object motion discontinuities, and other erroneous components of the input optic flow (errors from brightness constancy assumption etc.). This allows learning of a sparse egomotion basis set that is robust and useful for egomotion prediction under various types of uncertainties. In the following, we consider an information theory based perspective of the MFG network to gain insight on the robust representation by the hidden layer neurons, with a preference for sparse activations [1].

Suppose, $v$, $\hat{v}$, and $z$ represent the motion field (MF), input optic flow, and the hidden layer activations, respectively for the egomotion prediction setup. $\hat{v}$ is distorted by depth variations, object motion, and other noise sources. Although, the MFG network predicts rotational MF and unit depth translational MF separately, we merge them to a single MF term $v$ and its prediction by MFG as $\tilde{v}$, for notational simplicity. Therefore, we can rewrite Equation 6 in the paper as

$$\tilde{v} = MFG(\hat{v}) \qquad (1)$$

where, $\tilde{v}$ approximates $v$ and all of $v$, $\tilde{v}$ and $\hat{v}$ are $d$ dimensional real vectors.

Let, $v$, $\hat{v}$, and $z$ are samples drawn from random variables $V$, $\hat{V}$, and $Z$, respectively. $V \sim q(V)$ is the true generator process for $V$ and $\hat{V} \sim q_D(V)$ is the distortion process. The parametric encoder of MFG predicts latent representation $Z = f_\theta(\hat{V})$, which is deterministic for a parameter set $\theta$.

Since MFG is trained to predict $V$ using a sparse latent representation $Z$, the optimization will try to capture in $Z$ as much information about $V$ as possible, while also displaying sparse activations. Formally, the optimization will try to solve the following objective:

$$\operatorname*{argmax}_{\theta} I(V; Z) + \lambda \Psi(Z) \qquad (2)$$

where, $I(V; Z)$ is the mutual information between $V$ and the learned latent representation $Z$. $\Psi(Z) = 1/\sum(Z > 0)$ is the sparsity of activations and $\lambda$ is a user defined non-negative trade off between these two terms. The $\Psi(Z)$ function can be replaced with a suitable sparsity function such as the generalized logistic activation on Equation 9 of the paper.

$I(V; Z)$ can be written in terms of entropy of the distributions as, $I(V; Z) = H(V) - H(V|Z)$, where $H(V)$ is marginal entropy of $V$ and $H(V|Z)$ is the conditional entropy of $V$ given $Z$. Assuming the unknown MF distribution $q(V)$ to be constant, we can rewrite the objective function as

$$\operatorname*{argmax}_{\theta} -H(V|Z) + \lambda \Psi(Z) \qquad (3)$$

For the optimal $\theta$, conditional probability density $q(V|Z)$, and joint probability density $q(V, Z)$, we can write,

$$\max_{\theta} -H(V|Z) + \lambda \Psi(Z) = \max_{\theta} \mathbb{E}_{q(V,Z)}(\log q(V|Z))$$
$$+ \lambda \Psi(Z)$$
$$= \max_{\theta, p^*} \mathbb{E}_{q(V,Z)}(\log p^*(V|Z))$$
$$+ \lambda \Psi(Z)$$

where, $\mathbb{E}_{p(X)}(f(x)) = \int p(x) f(x) dx$ is expectation and we optimize over all possible distributions $p^*$. Expressing $p^*(V|Z)$ in a parametric form $p(V|Z)$ with parameter $\theta'$ (the decoder), we get the following lower bound.

$$\max_{\theta} -H(V|Z) + \lambda \Psi(Z) \geq \max_{\theta, \theta'} \mathbb{E}_{q(V,Z)}(\log p(V|Z))$$
$$+ \lambda \Psi(Z)$$

where, the equality is obtained for $p(V|Z) = q(V|Z)$. Since $\lambda \Psi(Z) \geq 0$, we can get rid of it from the expression.

$$\max_{\theta} -H(V|Z) + \lambda \Psi(Z) \geq \max_{\theta,\theta'} \mathbb{E}_{q(V,Z)}(\log p(V|Z))$$

Then, following Vincent et al. [1], considering $V$ to be a binary vector variable, we can express the reconstruction $p(V|Z) = B_{g_{\theta'}(Z)}(V)$ as a Bernoulli distribution with mean $g_{\theta'}(Z)$. Then the optimization over the lower bound can be expressed as,

$$\max_{\theta,\theta'} \mathbb{E}_{q(V,Z)}(\log B_{g_{\theta'}(Z)}(V))$$
$$= \max_{\theta,\theta'} \mathbb{E}_{q(V,\hat{V})}(\log B_{g_{\theta'}(f_\theta(\hat{V}))}(V))$$
$$= \min_{\theta,\theta'} \mathbb{E}_{q(V,\hat{V})}(L_H(V, g_{\theta'}(f_\theta(\hat{V}))))$$

where, $L_H(v, \tilde{v})$ is a negative log-likelihood for vector $v$, given the Bernoulli parameters $\tilde{v} = g_{\theta'}(f_\theta(v))$.

Therefore, minimizing the expected reconstruction loss using MFG is equivalent to maximizing a lower bound on the composite mutual information $I(V;Z)$, i.e between the true MF and its latent representation, and the sparsity of hidden layer activations $\Psi(Z)$.

## 2. Additional results

### 2.1. Object motion perception

We present additional object motion perception results obtained using our model on VKITTI object motion test split in Figure 1. The results show object motion extraction on a variety of egomotion and object motion conditions. Figure 1(i, v) depict extraction of a distant moving car with a small projected velocity when the egomotion is large. Figure 1 (iii, vi, and vii) depict performance on extraction of velocity of distant cars with small projected velocity in presence of nearer cars with large projected velocity. Figure 1 iv shows results in absence of any moving objects. Figure 1 (ii, viii) depict filtering of static objects with large projected egomotion velocity components.

### 2.2. Egomotion prediction

Figure 2 depicts representation of egomotion by hidden layer neurons on the KITTI egomotion split test set. From the source frame t, egomotion representation on the backward direction ($t \to t-1$) and forward direction ($t \to t+1$) are shown. The same group of 1000 hidden layer neurons represent translation and rotation for both forward and backward egomotion. In the frame t=1080, the car is moving mostly forward, which is encoded mostly by the hidden layer neurons with representing forward translation for $t \to t+1$ egomotion and backward translation for $t \to t-1$ egomotion. In frame t=1100, the car is going forward and to the right, therefore the most active neurons are selective for forward translation and rightward rotation for $t \to t+1$ egomotion and selective for backward translation and leftward rotation for $t \to t-1$ egomotion. For frame t= 1120 where the car is mostly turning to the right, MF representation activates the neurons with rightward rotation selectivity (for $t \to t+1$ egomotion) and leftward rotation selectivity (for $t \to t-1$ egomotion). Interestingly, it can be seen that the forward and backward egomotion activate different set of neurons, implying a specific egomotion preference is learned by each of the hidden neurons.

Figure 3 shows similar egomotion representation by the hidden layer neurons on the VKITTI egomotion split test set.

### 2.3. Sparse egomotion representation

Figure 4 depicts the effect of using a subset of the total 1000 hidden layer neurons of the MFG network for the prediction of egomotion on test sets. Overall, the network learns egomotion representations that are more sparse when trained on the KITTI dataset than on the VKITTI dataset. This may be because the KITTI odometry split training set has 12 times more frames than the VKITTI egomotion split training dataset. Moreover, the KITTI dataset is augmented by the forward and backward optic flow input, whereas the VKITTI dataset is not.

As shown in Figure 4(a), the network prediction is not much affected while using top $k = 2\%$ (or more) of the hidden layer neurons for both Sequence 09 and 10 of the KITTI odometry split test set.

On the VKITTI egomotion split test set (Figure 4(b) and 4(c)), the MFG network performance is maintained using only $15\%$ of the hidden layer neurons.

### 2.4. Egomotion basis

Figure 5 depicts the translational MF basis learned by 100 of the total 1000 hidden layer neurons. These neurons are selected in the decreasing order of L2 norm. Figure 6 depicts the 100 translational basis MF shown in Figure 5 using Euler coordinates, which are obtained using Equation 11 of the paper. Similarly, Figure 7 and Figure 8 depict 100 rotational basis MF learned by MFG hidden layer neurons, in pixelwise velocity and in Euler coordinates, respectively.

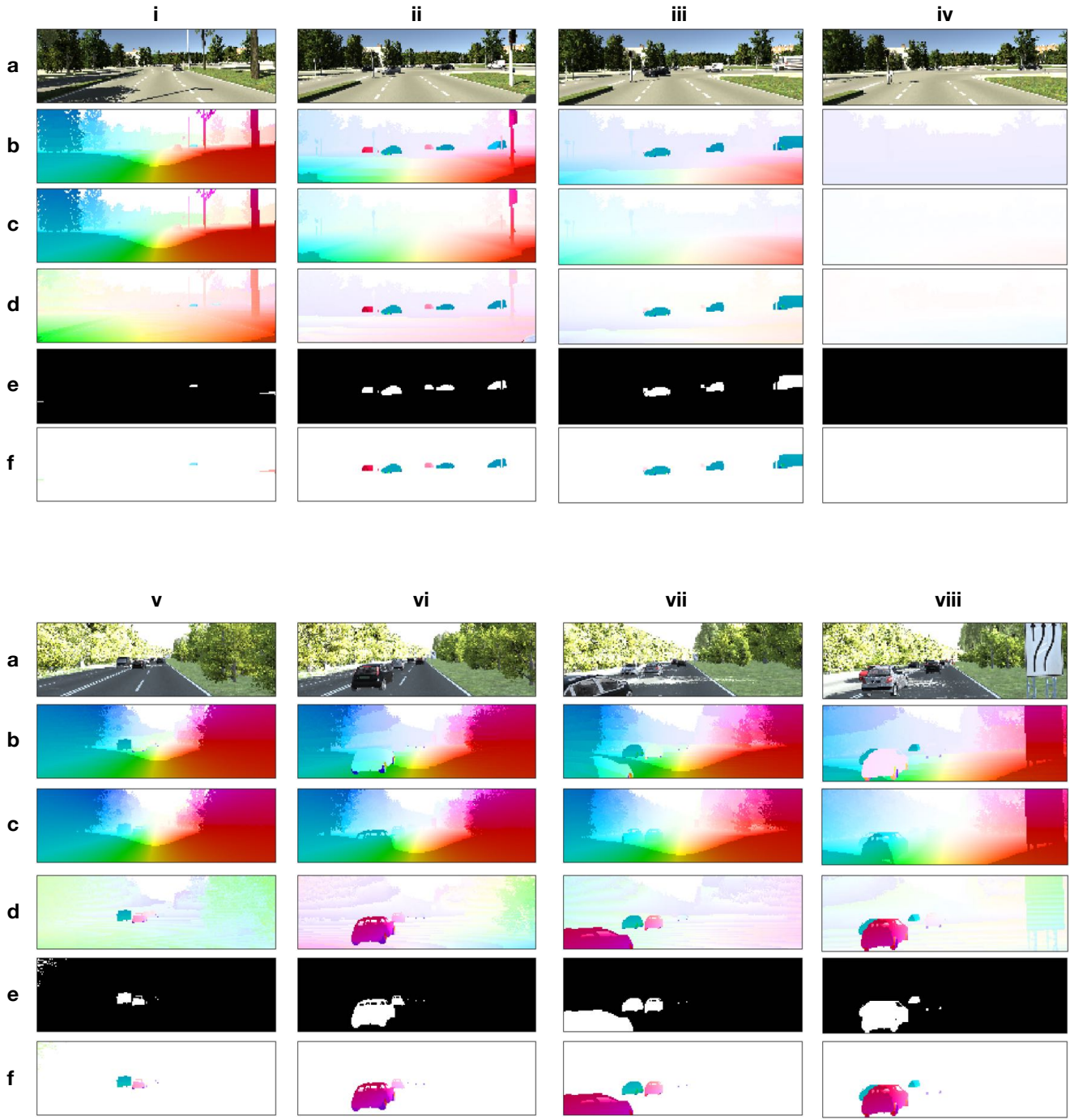

Figure 1. Object velocity prediction results on example sequences from VKITTI object-motion split. Frames i-iv are from Sequence 02 and frames v-viii are from Sequence 18. For each column i-viii, a is the RGB image, b is the optic flow, c is the predicted motion field, d is the predicted residue flow field, e is the predicted object motion mask, and f is the predicted object motion.

3

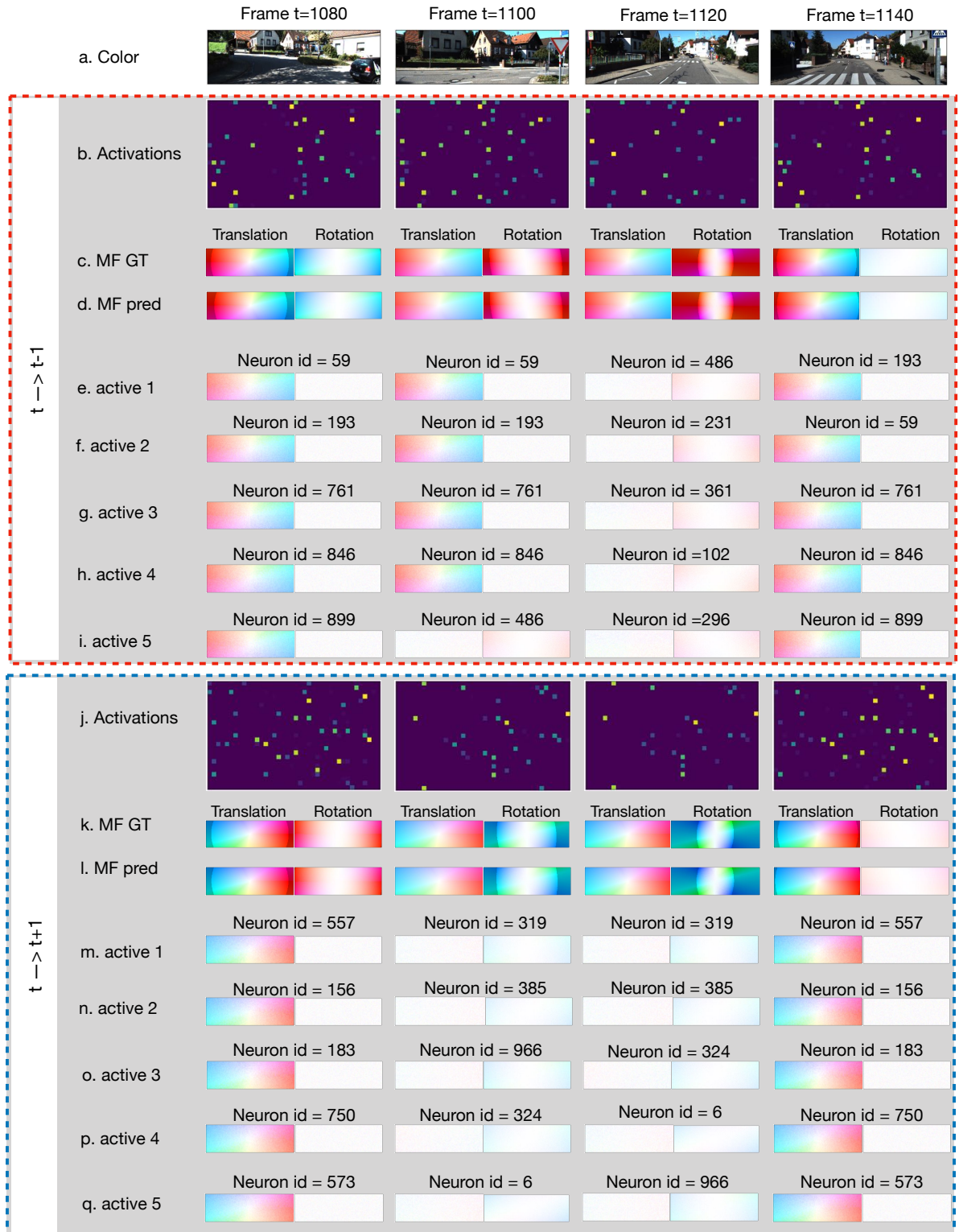

Figure 2. The MFG network representation of egomotion $t \to t-1$ (red dashed box) and $t \to t+1$ (blue dashed box) using sparsely activated basis features on KITTI odometry split test set. Each column shows the predicted motion fields and the activated basis components for a prediction. a) The source RGB frame, b,j) unit activation heat map of the hidden layer neurons (the brighter the color is, the more active the neuron is), c,k) GT translation-only (left, unit depth) and rotation-only (right) motion fields, d,l) predicted translation-only (left, unit depth) and rotation-only (right) motion fields, e-i, m-q) Translation and rotation flow fields represented by the five most active neurons in the hidden layer for each prediction.

4

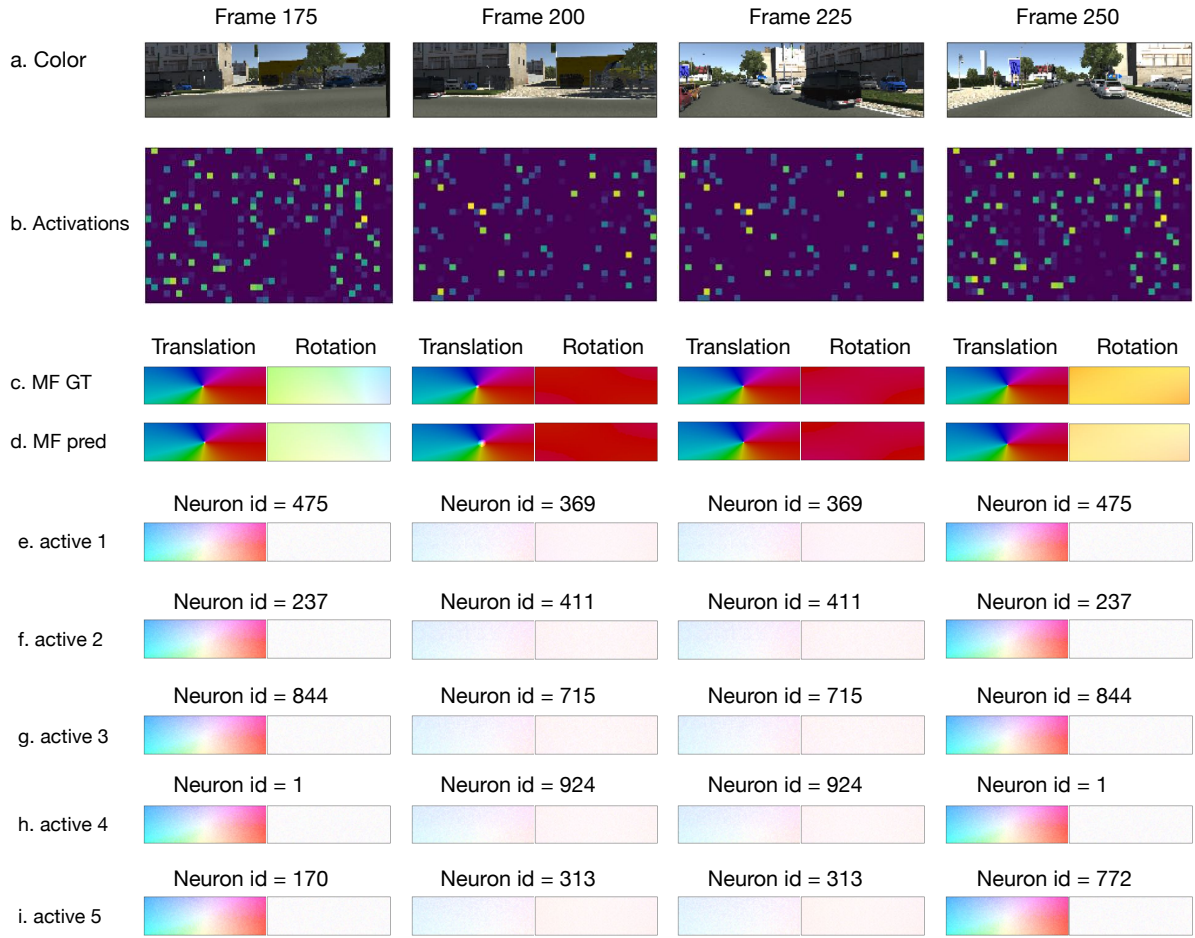

Figure 3. The MFG network representation of egomotion $t \rightarrow t+1$ using sparsely activated basis features on VKITTI egomotion split test set. Each column shows the predicted motion fields and the activated components for a source frame. a) The source RGB frame, b) unit activation heat map of the hidden layer neurons (the brighter the color is, the more active the neuron is), c) GT translation-only (left, unit depth) and rotation-only (right) motion fields, d) predicted translation-only (left, unit depth) and rotation-only (right) motion fields, e-i) Decoded translation and rotation flow fields by the five most active neurons in the hidden layer for each source frame.

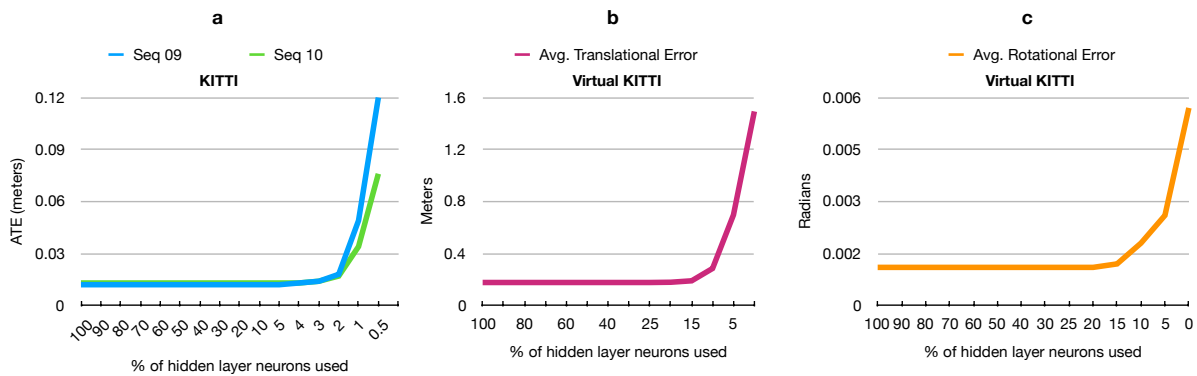

Figure 4. Performance of the MFG network for egomotion prediction against percentage of the hidden layer neurons are used to generate translation-only and rotation-only motion fields. a) ATE error on Sequence 09, 10 of the KITTI odometry split test set, b) Avg. translational error on Sequence 01 of VKITTI egomotion split test set, c) Avg. rotational error on Sequence 01 of VKITTI egomotion split test set.

5

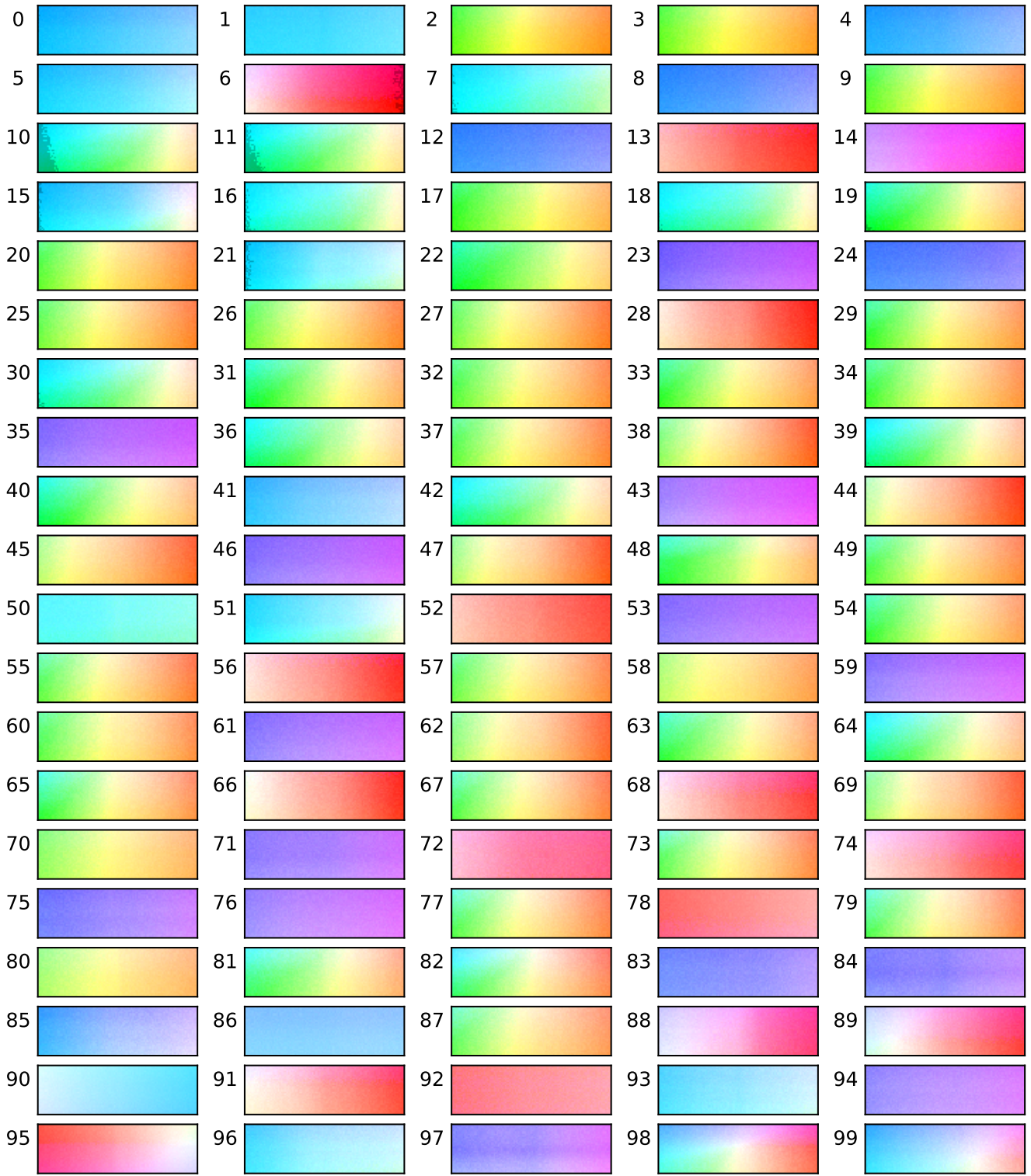

Figure 5. One hundred translational basis MF learned by MFG network when trained on the VKITTI egomotion split.

|    | X         | Y         | Z        |
|----|-----------|-----------|----------|
| 0  | -0.011309 | -0.001787 | 0.005436 |
| 1  | -0.011608 | -0.000053 | 0.002811 |
| 2  | 0.000107  | 0.009921  | 0.011075 |
| 3  | -0.000279 | 0.00996   | 0.010066 |
| 4  | -0.010099 | -0.003662 | 0.005144 |
| 5  | -0.010318 | -0.000423 | 0.006251 |
| 6  | 0.009629  | -0.000792 | 0.009089 |
| 7  | -0.009565 | 0.002566  | 0.007578 |
| 8  | -0.008708 | -0.005545 | 0.005018 |
| 9  | 0.00011   | 0.00892   | 0.010808 |
| 10 | -0.007292 | 0.004246  | 0.011567 |
| 11 | -0.007129 | 0.004745  | 0.011067 |
| 12 | -0.008036 | -0.005987 | 0.004539 |
| 13 | 0.009665  | 0.001169  | 0.005969 |
| 14 | 0.007588  | -0.005818 | 0.006129 |
| 15 | -0.008478 | 0.000134  | 0.010374 |
| 16 | -0.008142 | 0.00316   | 0.009586 |
| 17 | -0.002252 | 0.008298  | 0.009814 |
| 18 | -0.007776 | 0.003675  | 0.009452 |
| 19 | -0.004144 | 0.006601  | 0.011545 |
| 20 | 0.000687  | 0.007966  | 0.010883 |
| 21 | -0.008731 | 0.000749  | 0.008299 |
| 22 | -0.004736 | 0.006932  | 0.009657 |
| 23 | 0.001154  | -0.009156 | 0.00537  |
| 24 | -0.006458 | -0.006733 | 0.004457 |
| 25 | 0.001206  | 0.007643  | 0.010829 |
| 26 | 0.000853  | 0.007727  | 0.010688 |
| 27 | 0.00164   | 0.007492  | 0.010832 |
| 28 | 0.008066  | 0.002298  | 0.007946 |
| 29 | -0.000807 | 0.00717   | 0.011288 |
| 30 | -0.0064   | 0.003611  | 0.010946 |
| 31 | -0.003151 | 0.006388  | 0.011388 |
| 32 | 0.00023   | 0.007222  | 0.010806 |
| 33 | -0.001005 | 0.007013  | 0.011042 |
| 34 | -0.000433 | 0.007036  | 0.011069 |
| 35 | 0.001861  | -0.008459 | 0.004991 |
| 36 | -0.005157 | 0.004998  | 0.010411 |
| 37 | 0.000758  | 0.006969  | 0.010841 |
| 38 | 0.003224  | 0.005992  | 0.011272 |
| 39 | -0.004645 | 0.004627  | 0.011691 |
| 40 | -0.003511 | 0.005824  | 0.011138 |
| 41 | -0.008231 | -0.002006 | 0.005365 |
| 42 | -0.005625 | 0.004413  | 0.010273 |
| 43 | 0.003957  | -0.007373 | 0.005664 |
| 44 | 0.005431  | 0.004493  | 0.010393 |
| 45 | 0.003899  | 0.006212  | 0.009632 |
| 46 | 0.001087  | -0.008343 | 0.005249 |
| 47 | 0.003952  | 0.005776  | 0.010363 |
| 48 | -0.003028 | 0.006318  | 0.010292 |
| 49 | 0.00048   | 0.006552  | 0.01122  |
| 50 | -0.008352 | 0.002244  | 0.002604 |
| 51 | -0.007538 | 0.001105  | 0.00801  |
| 52 | 0.008067  | 0.001792  | 0.005151 |
| 53 | 0.000877  | -0.00833  | 0.004064 |
| 54 | -0.001097 | 0.006474  | 0.010702 |
| 55 | 0.001211  | 0.006142  | 0.011209 |
| 56 | 0.007565  | 0.000985  | 0.007098 |
| 57 | 0.000777  | 0.006508  | 0.010235 |
| 58 | 0.001573  | 0.007325  | 0.007458 |
| 59 | 0.001853  | -0.007749 | 0.005233 |
| 60 | 0.0003    | 0.00652   | 0.010096 |
| 61 | 0.001013  | -0.007807 | 0.004942 |
| 62 | 0.002861  | 0.005675  | 0.010198 |
| 63 | -0.001853 | 0.005846  | 0.010502 |
| 64 | -0.003965 | 0.004331  | 0.011026 |
| 65 | -0.000559 | 0.005552  | 0.011543 |
| 66 | 0.006737  | 0.001609  | 0.008356 |
| 67 | 0.000617  | 0.00573   | 0.011066 |
| 68 | 0.007292  | 0.000292  | 0.006834 |
| 69 | 0.003342  | 0.005128  | 0.01015  |
| 70 | -0.000167 | 0.007074  | 0.007436 |
| 71 | 0.000449  | -0.007627 | 0.004199 |
| 72 | 0.00768   | -0.000905 | 0.003527 |
| 73 | 0.000707  | 0.005718  | 0.010545 |
| 74 | 0.00695   | -0.000325 | 0.006453 |
| 75 | -0.001212 | -0.007178 | 0.004886 |
| 76 | 0.00195   | -0.007038 | 0.004674 |
| 77 | 0.000627  | 0.005198  | 0.010831 |
| 78 | 0.007509  | 0.000157  | -0.002794|
| 79 | 0.00053   | 0.00498   | 0.010806 |
| 80 | 0.000284  | 0.006583  | 0.006832 |
| 81 | -0.0023   | 0.00477   | 0.01017  |
| 82 | -0.000592 | 0.003904  | 0.012209 |
| 83 | -0.003571 | -0.006173 | 0.003468 |
| 84 | -0.002052 | -0.00678  | 0.003429 |
| 85 | -0.005211 | -0.003478 | 0.007054 |
| 86 | -0.006907 | -0.002165 | 0.001148 |
| 87 | 0.000498  | 0.005315  | 0.009501 |
| 88 | 0.005442  | -0.002002 | 0.008249 |
| 89 | 0.005311  | -0.000687 | 0.009283 |
| 90 | -0.006718 | -0.000157 | -0.004922|
| 91 | 0.005995  | 0.00079   | 0.007442 |
| 92 | 0.007117  | -0.000031 | -0.001749|
| 93 | -0.006616 | -0.000174 | 0.004931 |
| 94 | 0.000734  | -0.006601 | 0.004385 |
| 95 | 0.005706  | -0.000456 | -0.007842|
| 96 | -0.006419 | 0.000083  | 0.005269 |
| 97 | -0.00012  | -0.006231 | 0.005686 |
| 98 | 0.000152  | 0.000234  | 0.013366 |
| 99 | -0.003476 | -0.001255 | 0.011298 |

Figure 6. Preferred 3D translational motion selectivity in Euler coordinates of the one hundred basis MF shown in Figure 5.

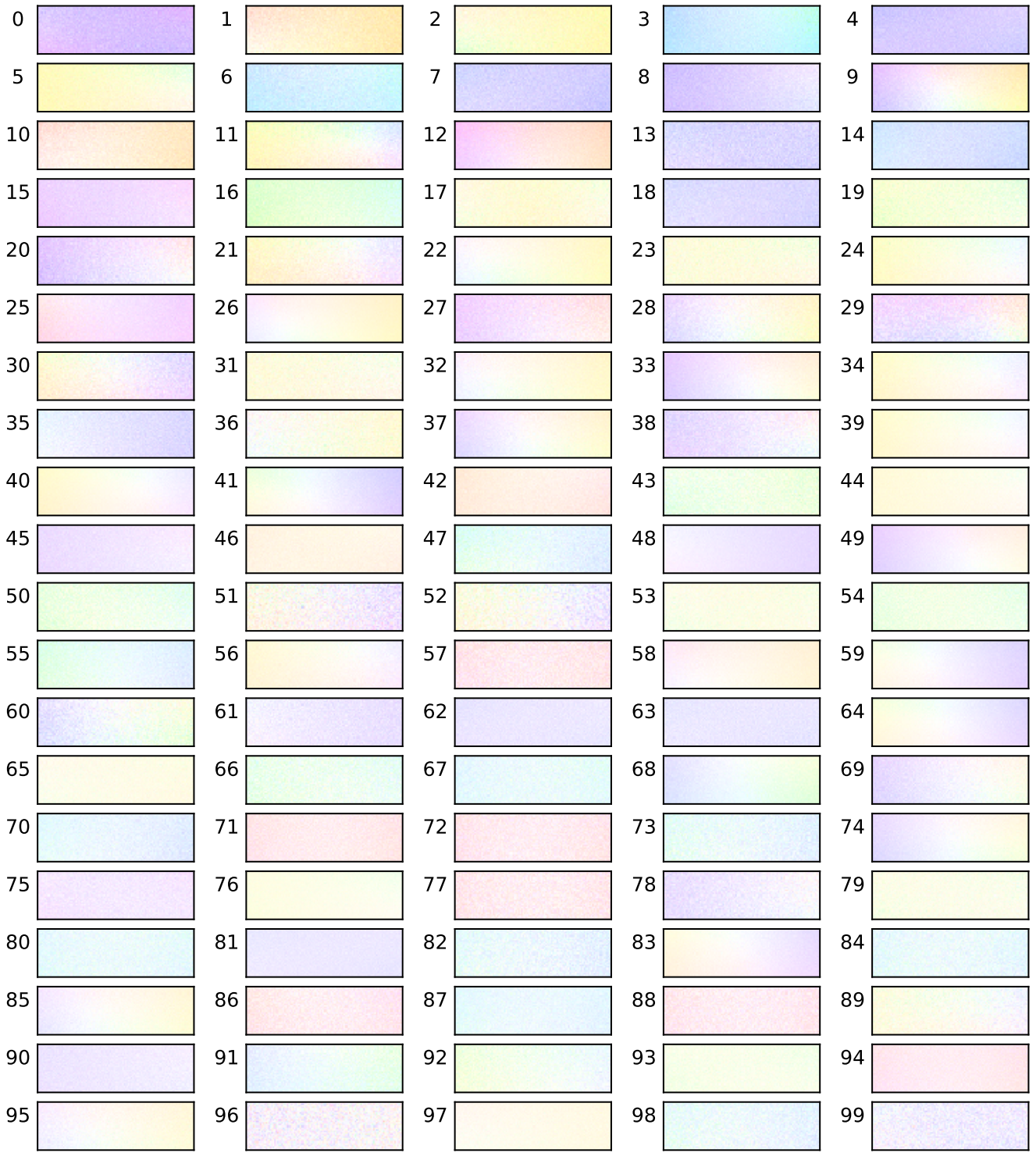

Figure 7. One hundred rotational basis MF learned by MFG network when trained on the VKITTI egomotion split.

|    | X         | Y         | Z         |
|----|-----------|-----------|-----------|
| 0  | -0.115958 | -0.013217 | -0.016008 |
| 1  | 0.095141  | -0.026359 | 0.068746  |
| 2  | 0.099099  | 0.00676   | 0.054502  |
| 3  | -0.008352 | 0.078977  | 0.047764  |
| 4  | -0.092757 | 0.015667  | 0.007837  |
| 5  | 0.085299  | 0.006717  | -0.060854 |
| 6  | -0.00998  | 0.067664  | 0.024861  |
| 7  | -0.076933 | 0.02577   | -0.017828 |
| 8  | -0.078062 | 0.006856  | 0.05158   |
| 9  | 0.014934  | -0.009902 | 0.153623  |
| 10 | 0.057974  | -0.032137 | 0.057751  |
| 11 | 0.053132  | 0.005433  | -0.102959 |
| 12 | 0.005803  | -0.047572 | 0.093163  |
| 13 | -0.0673   | 0.019456  | -0.009685 |
| 14 | -0.043584 | 0.046667  | -0.01852  |
| 15 | -0.060864 | -0.027233 | 0.024455  |
| 16 | 0.051021  | 0.036164  | -0.034277 |
| 17 | 0.066877  | 0.002283  | -0.003778 |
| 18 | -0.062745 | 0.020824  | -0.013857 |
| 19 | 0.060538  | 0.018657  | -0.026015 |
| 20 | -0.048881 | -0.013948 | 0.071546  |
| 21 | 0.035313  | -0.007379 | -0.094202 |
| 22 | 0.048969  | 0.002741  | 0.077354  |
| 23 | 0.059393  | 0.004743  | -0.01272  |
| 24 | 0.048566  | 0.007338  | -0.070651 |
| 25 | -0.034864 | -0.035744 | -0.037133 |
| 26 | 0.038649  | -0.007342 | 0.083309  |
| 27 | -0.029222 | -0.030321 | 0.056066  |
| 28 | 0.014455  | -0.000264 | 0.105222  |
| 29 | -0.02902  | -0.014282 | 0.067265  |
| 30 | 0.003749  | -0.008437 | -0.09864  |
| 31 | 0.050089  | 0.000334  | -0.023651 |
| 32 | 0.039216  | -0.001739 | 0.072311  |
| 33 | -0.014239 | -0.010378 | 0.099057  |
| 34 | 0.035157  | 0.00015   | -0.076448 |
| 35 | -0.041983 | 0.016823  | -0.039848 |
| 36 | 0.041101  | 0.001799  | 0.047832  |
| 37 | 0.00921   | -0.003942 | 0.09882   |
| 38 | -0.039786 | -0.00361  | 0.050067  |
| 39 | 0.038602  | 0.001485  | -0.065718 |
| 40 | 0.028772  | -0.004231 | -0.081359 |
| 41 | -0.016312 | 0.007687  | -0.091303 |
| 42 | 0.0306    | -0.028969 | -0.027288 |
| 43 | 0.038502  | 0.021331  | 0.014582  |
| 44 | 0.046085  | -0.001066 | -0.033237 |
| 45 | -0.044778 | -0.009908 | 0.025947  |
| 46 | 0.042259  | -0.015898 | -0.008088 |
| 47 | 0.001931  | 0.033264  | -0.032539 |
| 48 | -0.043651 | -0.003418 | -0.036483 |
| 49 | -0.01968  | -0.011089 | 0.080726  |
| 50 | 0.032782  | 0.022755  | -0.019059 |
| 51 | -0.001569 | -0.006716 | -0.060793 |
| 52 | 0.011773  | 0.000935  | -0.064052 |
| 53 | 0.043438  | 0.00622   | 0.005072  |
| 54 | 0.03595   | 0.020275  | -0.007506 |
| 55 | 0.00548   | 0.031056  | -0.03981  |
| 56 | 0.028417  | -0.006569 | -0.063279 |
| 57 | -0.000043 | -0.032563 | 0.004161  |
| 58 | 0.027599  | -0.014916 | 0.056079  |
| 59 | -0.018426 | 0.000269  | -0.078239 |
| 60 | 0.001902  | 0.012915  | 0.068348  |
| 61 | -0.038794 | 0.000411  | -0.025636 |
| 62 | -0.04172  | 0.006832  | 0.003819  |
| 63 | -0.040295 | 0.01025   | -0.002451 |
| 64 | -0.004508 | 0.002465  | -0.084676 |
| 65 | 0.041061  | 0.001257  | 0.010508  |
| 66 | 0.022424  | 0.024518  | -0.011548 |
| 67 | 0.002441  | 0.031451  | 0.007882  |
| 68 | 0.003465  | 0.020626  | 0.065344  |
| 69 | -0.018467 | -0.000171 | 0.069466  |
| 70 | -0.0116   | 0.029514  | -0.020861 |
| 71 | 0.002469  | -0.032011 | 0.003052  |
| 72 | -0.002275 | -0.030946 | 0.002904  |
| 73 | -0.000875 | 0.028178  | -0.021071 |
| 74 | -0.003983 | 0.001741  | 0.079859  |
| 75 | -0.034881 | -0.013469 | -0.000482 |
| 76 | 0.039225  | 0.005667  | -0.017851 |
| 77 | 0.000233  | -0.028955 | 0.000203  |
| 78 | -0.030911 | -0.00118  | 0.036748  |
| 79 | 0.037847  | 0.005899  | -0.005688 |
| 80 | 0.001621  | 0.029978  | 0.001098  |
| 81 | -0.03899  | 0.006656  | -0.003129 |
| 82 | -0.006513 | 0.02666   | -0.017142 |
| 83 | -0.002762 | -0.008201 | -0.076299 |
| 84 | 0.000058  | 0.028994  | -0.003391 |
| 85 | 0.016818  | -0.002157 | 0.070339  |
| 86 | -0.00112  | -0.028978 | -0.010038 |
| 87 | -0.003527 | 0.030008  | -0.008009 |
| 88 | -0.004022 | -0.028456 | -0.00208  |
| 89 | 0.024906  | 0.002975  | -0.046084 |
| 90 | -0.03718  | -0.000426 | 0.01495   |
| 91 | 0.005462  | 0.024964  | 0.036586  |
| 92 | 0.023222  | 0.015985  | -0.03833  |
| 93 | 0.035245  | 0.010951  | -0.003456 |
| 94 | -0.004963 | -0.030247 | 0.005834  |
| 95 | 0.02126   | -0.003178 | 0.059342  |
| 96 | -0.008082 | -0.006415 | 0.012753  |
| 97 | 0.035843  | -0.004039 | 0.01915   |
| 98 | -0.002309 | 0.024986  | -0.017544 |
| 99 | -0.025191 | 0.002435  | -0.010846 |

Figure 8. Preferred 3D rotational motion selectivity in Euler coordinates of the one hundred basis MF shown in Figure 7.



\end{document}

%% file: section_1.tex
The VO schemes to compute 6DoF camera motion based on motion of rigid features underperform in presence of independently moving objects~\cite{giachetti1998use,hyslop2010autonomous}. Particularly outdoor environments and a monocular setup create additional challenges to the existing online VO computation methods, due to noisy optic flow and depth information~\cite{srinivasan1994image}. Therefore, a robust method to compute the 6DoF egomotion parameters in dynamic outdoor scenes from noisy optic flow input is needed.

\begin{figure}[t]
\begin{center}
\includegraphics[width=0.75\linewidth]{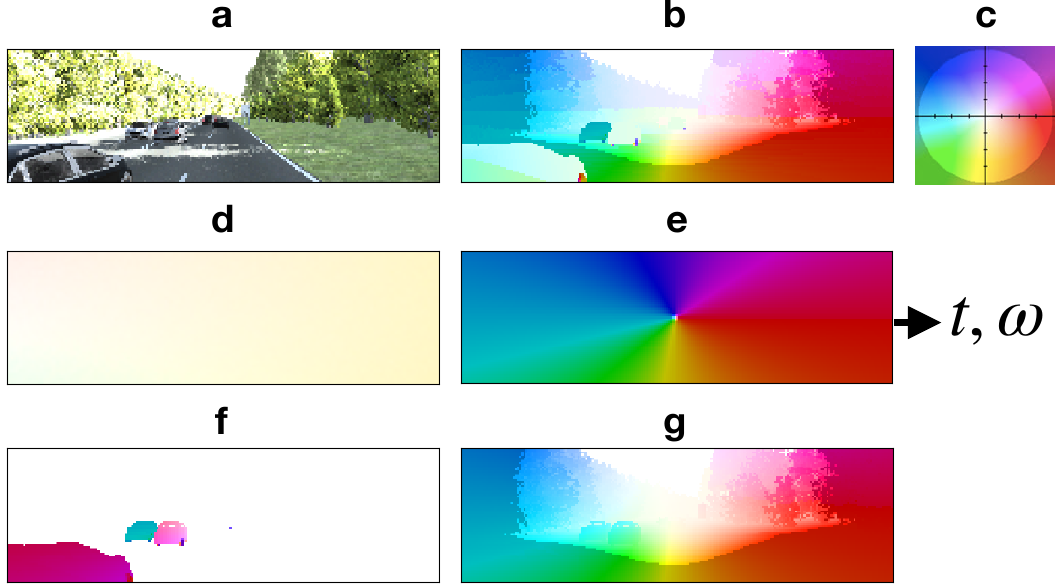}
\end{center}
\caption{Egomotion and object-motion predictions by the proposed model. a) RGB frame, b) optic flow, c) color coding as in~\cite{baker2011database} used for optic flow, motion field, and object motion throughout this paper, d,e) predicted rotational and translational egomotion fields that are converted to 6DoF camera motion parameters, g) predicted egomotion field with depth information, f) predicted object motion.}
\label{fig:demo}
\end{figure}

Not only is estimating egomotion parameters in a dynamic scene important, many applications such as autonomous navigation and tracking, require computation of velocities of the independently moving objects online while the observer is also moving. In order to compute object velocity, the observer's egomotion needs to be compensated~\cite{bak2014dynamic}. The state-of-the-art tracking by detection approaches cannot derive the actual velocity of the objects without egomotion estimate~\cite{kalal2012tracking,wang2013learning,henriques2015high}. Recent deep learning based structure-from-motion (SfM) methods do not estimate object velocity~\cite{zhou2017unsupervised,mahjourian2018unsupervised,godard2018digging}. Other similar approaches that distinguish static and dynamic segments do not explicitly separate egomotion and object motion velocity components for the dynamic regions~\cite{lee2018cemnet,yin2018geonet,yang2018every,vijayanarasimhan2017sfm}.

The goal of this work is to separate optic flow into image velocity components due to egomotion and object motion across the entire image. We propose a sparse autoencoder network that learns to predict the image velocity components due to camera translation and rotation from optic flow input, marginalizing scene depth and noise. The predicted egomotion components are then converted to 6DoF camera motion parameters in closed form using the geometry of instantaneous rigid body motion under perspective projection~\cite{bruss1983passive}. For object motion prediction, the predicted egomotion is first scaled by scene depth, which is then compensated from optic flow and processed through multi stage max-pooling normalization to compute object motion~\cite{carandini2012normalization}.

We train the neural network using a supervised loss from egomotion field constructed from GT pose information, such that the training conforms with the geometry of instantaneous rigid body motion~\cite{lee2018cemnet}. On the KITTI visual odometry dataset~\cite{geiger2012we}, the proposed model achieves the state-of-art performance for camera trajectory  prediction. On the Virtual KITTI dataset~\cite{gaidon2016virtual}, the proposed model achieves the state-of-the-art performance in terms of rotational error and comparable performance for translational error in the camera movement prediction task.

An additional advantage of the proposed approach over the existing methods is that the sparse autoencoder learns an overcomplete set of meaningful basis motion fields for camera translation and rotation when trained with a regularizer for sparsity. Sparse representation of data has many benefits, such as, i) eliminating irrelevant variabilities in input data and making the network robust to outliers, ii) finding hidden structures that are more suitable as input for machine learning applications due to increased separability, and iii) fewer active neurons, which leads to fewer downstream computations~\cite{boureau2008sparse,poultney2007efficient,ngiam2011sparse}. The proposed model learns a sparse egomotion representation that successfully marginalizes object velocities (outliers) and scene depth variations in the input optic flow data to learn basis motion fields related to camera translation and rotation.

The learned sparse egomotion representation generalizes well to test data and the network achieves good egomotion prediction accuracy in presence of independently moving objects and in novel scene structures. Further, only 5\% of the hidden layer neurons are required on the KITTI test set for predicting rotation and translation of the camera without any drop in performance. Finally, the learned egomotion representations by individual hidden neurons are meaningful in terms of their selectivity for a particular type of camera translation and rotation.

%% file: section_2.tex

\subsection*{Monocular continuous methods for VO}
Most of the existing methods for monocular continuous VO cannot handle noise or the presence of independently moving objects~\cite{zhang2002consistency}. Fredriksson et al. recently proposed branch and bound methods that are able to compute translational velocity in presence of outliers~\cite{fredriksson2015practical,fredriksson2014fast}. Jaegle et al. estimated general camera motion from noisy optic flow input using confidence score and an iterative optimization procedure on sparse optic flow~\cite{jaegle2016fast}. Lee and Fowlkes solved for dense optic flow for static and dynamic segments and then formulate for egomotion recovery using static flow segment and scene depth input~\cite{lee2018cemnet}. The proposed approach is related to this class of approaches, however is focused on direct egomotion prediction from optic flow by marginalizing outliers and scene structure variations using a sparse representation of egomotion components.

\subsection*{Structure-from-motion prediction}
The traditional Structure-from-motion(SfM) methods rely on accurate feature matching, which requires good photo consistency promise~\cite{schonberger2016structure,triggs1999bundle}. Recent development in deep neural networks have allowed formulation of the SfM as a prediction problem and many new methods have been proposed in the recent years on this theme. Zhou et al. proposed a network that learns to predict depth and camera motion and is trained using self supervised image warping loss between the source and the target frames~\cite{zhou2017unsupervised}. This self supervised training has since been adopted by many other approaches~\cite{mahjourian2018unsupervised,godard2018digging,yin2018geonet,yang2018every}. Vijayanarasimhan et al. added a separate network to predict a predetermined number of object layers during camera movement~\cite{vijayanarasimhan2017sfm}. However, their training procedure does not explicitly separate individual velocity components from egomotion and object motion. Tung et al. formulate the same problem in an adversarial framework where the generator is trained to synthesize camera motion and scene structure that minimize the warping cost to a target frame~\cite{tung2017adversarial}.
%
%

\subsection*{Sparse representation}
Sparse representation was initially used as a preprocessing step for classification tasks, because sparse features were found to be more separable than the input data itself~\cite{ngiam2011sparse}. Since a small population of features are used to represent the variability among large set of data points, these sparse representation methods were found to be learning hidden structures common across a large subset of data and ignoring variabilities inconsistent across datapoints, i.e. noise~\cite{poultney2007efficient,boureau2008sparse}. Applications that use the learned sparse representation of data benefit from sparsity by performing fewer computations. So far, multiple schemes of learning sparse representations have been proposed, such as sparse coding~\cite{lee1999learning}, sparse autoencoder~\cite{coates2011analysis}, sparse winner-take-all circuits~\cite{ngiam2011sparse,makhzani2015winner} and sparse RBMs~\cite{lee2008sparse} for learning components of faces, handwritten digits, edges in natural scenes etc.

Compared to these approaches, our data is sequential and highly skewed towards some particular types of motion. Therefore, some of the common sparsity constructs, such as lifetime sparsity~\cite{ngiam2011sparse}, do not apply to our case.

%% file: section_3.tex

\section{Continuous egomotion formulation}
The continuous egomotion formulation has been used by previous methods to estimate camera motion and/or rigid scene structure from optic flow of monocular visual input~\cite{bruss1983passive,jepson1991fast,zhang1999fast}. Given the instantaneous camera translation velocity $t=(t_x, t_y, t_z)^T\in R^3$, the instantaneous camera rotation velocity $\omega \in (\omega_x, \omega_y, \omega_z)^T \in R^3$, and the inverse of scene depth $\rho (p_i) = \frac{1}{Z(p_i)} \in R$, the image velocity $v(p_i)=(v_i, u_i)^T \in R^2$ at an image location $p_i=(x_i, y_i)^T \in R^2$ of an calibrated camera image due to camera motion is given by,

\begin{equation}
v(p_i) = \rho (p_i) A(p_i)t + B(p_i)\omega
\end{equation}

where,

\begin{equation*}
A(p_i) = \begin{bmatrix}
f & 0 & -x_i \\
0 & f & -y_i
\end{bmatrix}
\end{equation*}

\begin{equation*}
B(p_i) = \begin{bmatrix}
-x_iy_i & f+x_i^2 & -y_i \\
-f-y_i^2 & x_iy_i & x_i
\end{bmatrix}
\end{equation*}

If $p_i$ is normalized by the focal length $f$, then it is possible to replace $f$ with $1$ in the expressions for $A(p_i)$ and $B(p_i)$.

If the image size is N pixels, then the full expression of instantaneous velocity at all the points, termed as motion field (MF), can be expressed in a compressed form as

\begin{equation}
v = \rho At + B\omega \label{eq:mf}
\end{equation}

where, $A$, $B$, and $\rho$ entails the expressions $A(p_i)$, $B(p_i)$, and $\rho (p_i)$ respectively for all the N points in the image as follows.

\begin{equation*}
v = \begin{bmatrix}
v(p_1) \\
v(p_2) \\
\vdots \\
v(p_N)
\end{bmatrix} \in R^{2N\times 1},\quad 
\rho At = \begin{bmatrix}
\rho_1 A(p_1) t \\
\rho_2 A(p_2) t \\
\vdots \\
\rho_N A(p_N) t
\end{bmatrix}\in R^{2N\times 1}
\end{equation*}

\begin{equation*}
B\omega = \begin{bmatrix}
B(p_1) \omega \\
B(p_2) \omega \\
\vdots \\
B(p_N) \omega
\end{bmatrix}\in R^{2N\times 1}
\end{equation*}

The monocular visual egomotion computation uses this formulation to estimate the unknown parameters $t$ and $\omega$ and the $N$ values of depth given the point velocities $v$ generated by camera motion. However, instantaneous image velocities obtained from the standard optic flow methods on real data are usually different from the real pixel velocities due to camera motion and scene structure, known as motion field (MF)~\cite{lee2018cemnet}. The presence of moving objects further deviates the optic flow away from the true MF. Let us call the input optic flow as $\hat{v}$, which is different from $v$. Therefore, monocular continuous methods on real data solve the following minimization objective to find $t$, $\omega$, and $\rho$.

\begin{equation}
t^*,\omega^*,\rho^* = \argmin_{t, \omega, \rho} \norm{\rho At + B\omega - \hat{v}}^2
\end{equation}

As found in~\cite{zhang1999fast,jaegle2016fast}, without loss of generality, the objective function can be first minimized for $\rho$ as,

\begin{align}
t^*,\omega^*,\rho^* &= \argmin_{t, \omega} \argmin_{\rho} \norm{\rho At + B\omega - \hat{v}}^2 \\
t^*,\omega^* &= \argmin_{t, \omega} \norm{{A^{\perp}t}^T (B\omega - \hat{v})}^2 \label{eq:cem}
\end{align}

where ${A^{\perp}t}$ is orthogonal complement of ${At}$. This resulting expression does not depend on $\rho$ and can be optimized directly to find optimal $t^*$ and $\omega^*$. 



%% file: section_4.tex
Herein, we propose Motion Field Generator (MFG), a neural network model, that learns to predict egomotion fields due to rotation and translation from optic flow input of dynamic scenes. These are converted to 6DoF egomotion parameters in closed form using constructs of rigid body motion. A secondary multi-stage max-pooling normalization module extracts the object motion after compensating for egomotion field scaled by scene depth.
We train the MFG network using ground truth pose information and a sparsity penalty, to learn meaningful representations of egomotion that are sparsely activated and robust to noise in the input optic flow.

\begin{figure}[h]
\begin{center}
\includegraphics[width=0.9\linewidth]{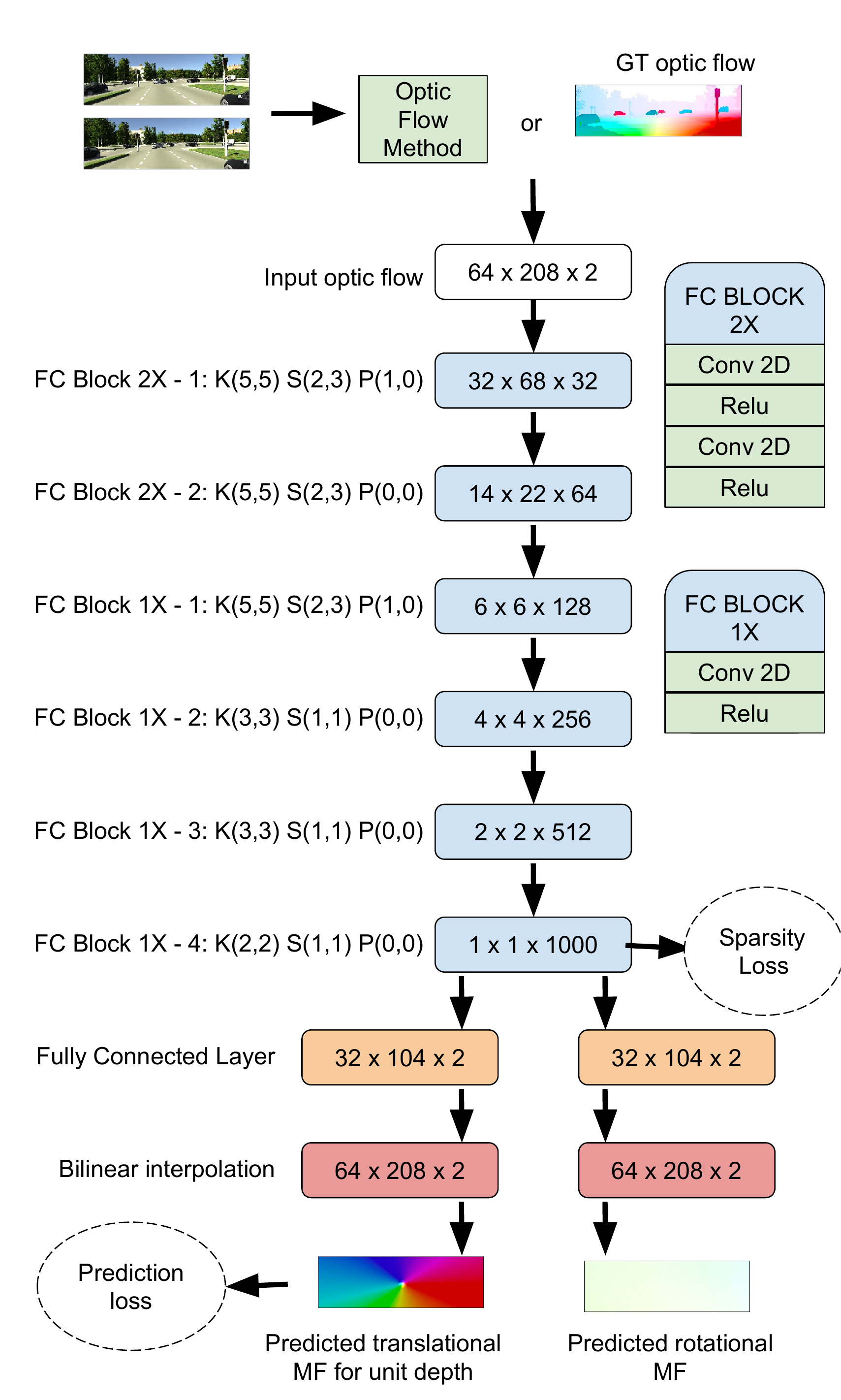}
\end{center}
\caption{Architecture of the proposed MFG network. FC Blocks are fully convolutional layers of 2D convolution and relu operations. The receptive field size is gradually increased such that the FC Block 1X-4 layer operates across the entire image. Output of all FC layers are non-negative due to relu operations. K, S, and P denote the kernel sizes, strides, and padding along vertical and horizontal directions of feature maps. The weights of the fully connected layer forms the basis for translational and rotational egomotion.}
\label{fig:mfg}
\end{figure}

Figure~\ref{fig:mfg} depicts the architecture of the proposed MFG network. The network is an asymmetric autoencoder that has a multilayer fully convolutional (FC) encoder and a single layer linear decoder.
The output neurons of the FC Block 1X-4 layer at the end of the encoder learns a latent space representation of egomotion. We will refer to this layer of $M=1000$ neurons as the hidden layer of the MFG network. The activations of all FC layer outputs in the encoder, including the hidden layer neurons, are non-negative due to relu operations.

\subsection*{Egomotion prediction loss}

For an optic flow input $\hat{v}$, the MFG network generates a rotational MF prediction$\tilde{v_\omega}$ and a depth marginalized translational MF prediction $\tilde{v_t}$.

\begin{equation}
\tilde{v_\omega},\quad \tilde{v_t} = MFG(\hat{v})
\end{equation}




We can obtain the translational prediction error and the rotation prediction error as,

\begin{equation}
L_t = \sum_p{\norm{v_t - \tilde{v_t}}^1}, \quad L_\omega = \sum_p{\norm{v_\omega - \tilde{v_\omega}}^1}
\end{equation}


where, $v_t$ is GT translational MF with $\rho=1$ and $v_\omega$ is GT rotational MF, obtained using Equation~\ref{eq:mf}.

We use L1 norm to calculate prediction error to avoid biases from large outliers.

Between rotation and translation, some datasets disproportionately represent one type of egomotion over the other, such as $\norm{v_w}$ is often small compared to $\norm{v_t}$ averaged over a real dataset resulting in significantly smaller rotational losses. In order to remove such biases and let the MFG network learn representations of both translation and rotation equally, we normalize $L_t$ and $L_\omega$ by scaling them using the following coefficients:

\begin{equation}
w_t = max(\frac{\norm{v_\omega}^2}{\norm{v_t}^2}, 1), \quad w_\omega = max(\frac{\norm{v_t}^2}{\norm{v_\omega}^2}, 1)
\end{equation}


\subsection*{Sparsity loss}

The MFG network is regularized during training for sparsity of activation of the hidden layer neurons. This is implemented by calculating a sparsity loss ($L_s$) for each batch of data and backpropagating it along with the egomotion prediction loss during training. The value of $L_s$ is calculated for each batch of data as the number of non-zero activations of the hidden layer neurons, also known as population sparsity. Although, to make this loss differentiable, we approximate a number of activations using a generalized logistic activation $g$ with a sharp saturation. The expression for $L_s$ is given as,

\begin{equation}
L_s = \sum_{i=1}^{M}{g(h_i)}
\end{equation}

where $h_i$ is the instantaneous activation of the $i$-th hidden layer neuron and $g(h_i)$ is given by,

\begin{equation*}
g(h_i) = \frac{1}{1 + Q e^{-B h_i}}
\end{equation*}

The coefficients $Q$ and $B$ are set to 25 and 10, respectively to approximate the ``$>0$" function.

Therefore, the total loss for training the MFG is given by,

\begin{equation}
L = w_tL_t + w_\omega L_\omega + w_sL_s 
\end{equation}

where, $w_s$ is a user selected hyperparameter for the coefficient of sparsity loss. It affects the sparsity of the representation learned by MFG for egomotion prediction.

\subsection*{Egomotion parameter estimation}
The predicted translation and rotation parameters can be computed directly from $\tilde{v}_t$ and $\tilde{v}_\omega$ following the continuous egomotion formulation.
\begin{equation}
\tilde{t} = \tilde{v}_t/A \mid \rho=1, \quad \tilde{\omega} = \tilde{v}_\omega/B
\end{equation}

\subsection*{Object motion extraction}
Scene depth $d$ (or its inverse$\rho$) is required to compute object velocity after compensating egomotion field from optic flow. The residual image velocity $r = \hat{v} - \tilde{v}$ calculated using the predicted MF as $\tilde{v} = \rho \tilde{v_t} + \tilde{v_\omega}$ results in additional residue other than object velocity, we call them residue noise. They may be due to i) inaccurate prediction by MFG $\tilde{v} \ne v$, ii) noisy input optic flow $\hat{v} \ne v+ v_{obj}$, and/or iii)inaccurate depth map or $\rho$ etc. We observe that nearby static objects cause large residue noise. A simple thresholding operation fails since that removes slowly moving objects.
\begin{figure}[h]
\begin{center}
\includegraphics[width=0.4\linewidth]{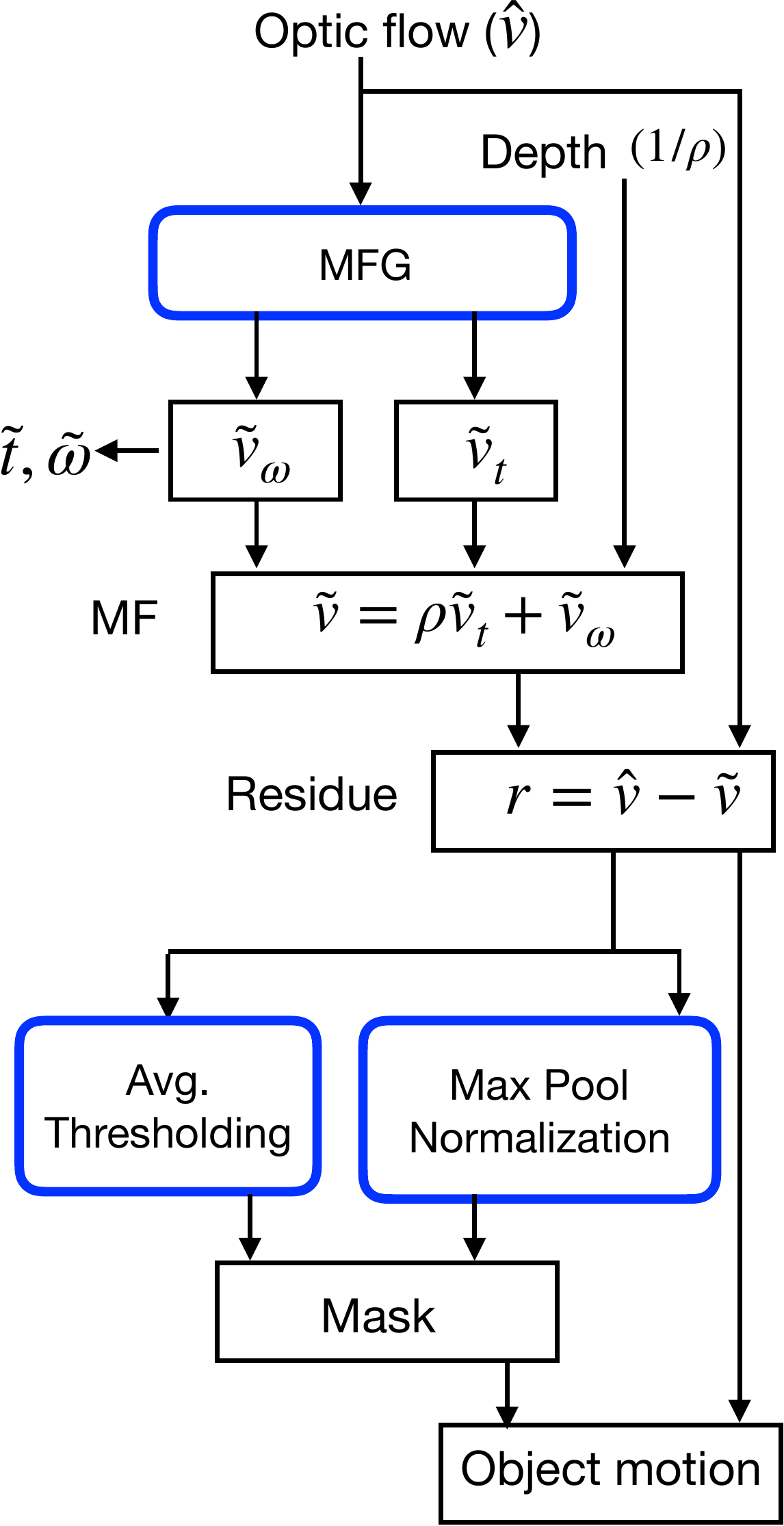}
\end{center}
\caption{Computational graph for egomotion and object-motion computation.}
\label{fig:object-arch}
\end{figure}

In order to find masks for only the moving objects in $r$, we employ the conjunction of two types of filters, discussed below:

\subsubsection*{Filter 1: Average thresholding}
The first filter is simple thresholding based on average, which keeps the objects as well as large residues from static parts.

The first filter $F_1$ is derived as,
\begin{equation*}
F_1 = \norm{r}^2 > \theta_d \times avg (tanh(\norm{r}^2))
\end{equation*}
where $\theta_d$ is a constant parameter.

\subsubsection*{Filter 2: Multistage max pooling}

The second filter is a multistage divisive max-pooling normalization scheme based on depth marginalized residue values. This filter assigns high probability to independent object movements at all depths. however, due to depth marginalization, the distant residues get multiplicatively scaled to large values.

In order to calculate the second filter $F_2$, we first normalize the depth $d=\frac{1}{\rho}$ to range 0-1. Then we perform non maximal suppression on $r$ by performing the following steps,
\begin{align*}
1)&\quad r' = \norm{r}^2 / max(\norm{r}^2), \quad
2)\quad r'' = r' \circ {tanh(d)}^2 \\
3)& r'' = r''/max(r''),
4) r_s = r'' \circ {(1 + (max(r'') - r'')^2)}^{-1}\\
F_2 &= (r_s > (\theta_p \times avg(r_s))) \circ (d<Threshold_D)
\end{align*}
From $F_1$ and $F_2$, the binary object motion mask is calculated as
\begin{equation*}
F_{obj} = F_1 \circ F_2
\end{equation*}
This object motion mask is then multiplied pixelwise with the residue to find object velocity.
\begin{equation}
\tilde{v}_{obj} = r \circ F_{obj}
\end{equation}

%% file: section_5.tex
We evaluate the performance of the proposed MFG network model in i) egomotion prediction from optic flow and ii) object motion extraction tasks.
Further, we analyze the latent space representation learned by the proposed model for rotational and translational basis flow fields and the sparsity of egomotion representation on unseen data.

\subsection{Datasets}
\subsubsection*{KITTI visual odometry split}
\label{sec:kitti_vis_odom}
We use the KITTI visual odometry dataset~\cite{geiger2012we} to evaluate egomotion prediction performance by the proposed model. This dataset provides eleven driving sequences (00-10) with RGB frames (we use only the left camera frames) and the ground truth pose for each frame. Of these eleven sequences, we use sequences 00-08 for training our model and sequences 09, 10 for testing, similar to Zhou et al.~\cite{zhou2017unsupervised}. This amounts to approximately 20.4K frames in the training set and 2792 frames in the test set. As ground truth optic flow is not available for this dataset, we use a pretrained PWC-Net~\cite{sun2018pwc} model to generate optic flow from the pairs of consecutive RGB frames for both training and testing.

\subsubsection*{Virtual KITTI split }
We use the Virtual KITTI dataset~\cite{gaidon2016virtual} to evaluate egomotion prediction performance, as well as object-motion extraction. The dataset provides ground truth optic flow, depth, and pose for five driving sequences that clone the real KITTI scenes and camera viewpoints~\cite{geiger2012we}. The advantages of Virtual KITTI dataset are precise ground truth camera motion than the KITTI dataset for short subsequences~\cite{tung2017adversarial} and the presence of more moving objects per frame than KITTI. For the egomotion prediction task from optic flow, we use the first sequence as the test set and the rest four sequences as the training set, for direct comparison with the existing methods on this dataset~\cite{tung2017adversarial,lee2018cemnet}. This will be referred to as VKITTI egomotion split. However, the first sequence has comparatively few frames with large components of both egomotion and object motion. In order to test object motion perception for a wide range of egomotion, we use sequences 0002 and 0018 for testing and the sequences 0001, 0006, and 0020 for training the MFG network. We will refer to this as VKITTI object-motion split.


\subsection{Training}
All the models are trained and tested using PyTorch 0.4.1 on a dual GeForce 960 GPU based system or a single 1080Ti GPU based system. We use Adam optimizer~\cite{kingma2014adam} and the same set of training hyper-parameters to train the MFG network on all training sets. A fixed learning rate $\eta$ of $10^{-5}$ is used in all experiments. We observe that full stochastic optimization, i.e. batch size 1 with random shuffling, paired with high values of the forgetting factors for gradients and second moment of gradients (the $\beta$ parameters of Adam) results in better generalization to the test set. Similar observation was made for training Fully Convolutional Networks earlier~\cite{shelhamer2016fully}, where they use stochastic gradient descent optimization with batch size 1 and momentum 0.99. This mode of parameter search helps to come out of local minima, while still avoiding deviations caused by outliers. Therefore, we use batch size 1 with random shuffling and $\beta_1 = 0.99$ and $\beta_2 = 0.999$ in all experiments. The sparsity coefficient $\alpha_s$ for training is set to $10^2$.

\subsection{Evaluation method for sparsity}
We study the sparsity of the representation learned by the MFG network by measuring the performance of the network for egomotion prediction on the test set using as few hidden layer neurons as possible. In order to measure the sparsity of the representation learned, during test, we find the top $k\%$ active neurons in the hidden layer for each input frame sequence and set the rest of the hidden layer activations to zero. Then we observe how the performance of the network degrades as we lower the value of $k$. Similar evaluation of sparsity was used in~\cite{makhzani2015winner}. The training is still done the same way with the sparsity penalty added to the error term for backpropagation, and without any hard reset of the hidden layer neurons, unlike~\cite{makhzani2015winner}.
This is to find a latent space during optimization that naturally represents egomotion using a sparse set of components.

\subsection{Egomotion prediction}

\subsubsection*{KITTI}
Following the egomotion evaluation protocol by Zhou et al.~\cite{zhou2017unsupervised} and others, each driving sequence of the training and test data of the KITTI visual odometry split is divided into overlapping snippets of five consecutive frames. The sequence length is set to be five during training. Absolute Trajectory Error (ATE) is used as the metric for evaluation, which measures the difference between the corresponding points of the ground truth and the predicted trajectories. For each test sequence, ATE is computed individually for each of the five frame snippets and then averaged over all the snippets in the sequence. In Table~\ref{tab:kitti-ego}, we compare the proposed model against the existing methods that use the same input setting for egomotion evaluation on the KITTI odometry split. For reference, we also compare against a traditional SLAM method ORB-SLAM~\cite{mur2015orb} that receives the whole sequence as input, referred to as ORB-SLAM(full). The performance of the same method using the same five frame snippet input setting as ours is referred to as ORB-SLAM(short).

\begin{table}[h]
\begin{tabular}{lcc}
\hline
Method           & Seq 09 & Seq 10 \\
\hline
\hline
ORB-SLAM (full)~\cite{mur2015orb} & 0.014$\pm$0.008  & \textbf{0.012$\pm$0.011}  \\
\hline
ORB-SLAM (short)~\cite{mur2015orb}* & 0.064$\pm$0.141  & 0.064$\pm$0.130  \\
Zhou et al.~\cite{zhou2017unsupervised}*      & 0.021$\pm$0.017  & 0.020$\pm$0.015  \\
Lee and Fowlkes~\cite{lee2018cemnet}$\dagger$       & 0.019$\pm$0.014  & 0.018$\pm$0.013  \\
Yin et al.~\cite{yin2018geonet}$\ddagger$ & \textbf{0.012$\pm$0.007}& \textbf{0.012$\pm$0.009}\\
Mahjourian et al.~\cite{mahjourian2018unsupervised}$\otimes$ & 0.013$\pm$0.010& \textbf{0.012$\pm$0.011}\\
Godard et al.~\cite{godard2018digging}$\dagger$ & 0.023$\pm$0.013& 0.018$\pm$0.014\\
Ours* & 0.020$\pm$0.015 & 0.023$\pm$0.015\\
Ours$\dagger$ & \textbf{0.012$\pm$0.006}& 0.013$\pm$0.008\\
Ours (Sparse $k=10\%$)$\dagger$ & \textbf{0.012$\pm$0.006}& 0.013$\pm$0.008\\
Ours (Sparse $k=5\%$)$\dagger$ & \textbf{0.012$\pm$0.006}& 0.013$\pm$0.008\\
Ours (Sparse $k=4\%$)$\dagger$ & 0.013$\pm$0.006& 0.013$\pm$0.008\\
Ours (Sparse $k=3\%$)$\dagger$ & 0.014$\pm$0.007& 0.014$\pm$0.008\\
Ours (Sparse $k=2\%$)$\dagger$ & 0.018$\pm$0.008& 0.017$\pm$0.010\\
Ours (Sparse $k=1\%$)$\dagger$ & 0.049$\pm$0.066& 0.034$\pm$0.046\\
Ours (Sparse $k=0.5\%$)$\dagger$ & 0.120$\pm$0.170& 0.076$\pm$0.120\\
\hline \\
\end{tabular}
\caption{ATE on the KITTI visual odometry split averaged over all five frame snippets (except for Mahjourian et al.~\cite{mahjourian2018unsupervised} three frame snippets). * The sequence length is equal to five during test. $\ddagger$ The sequence length during test is unknown. $\dagger$ Frame-by-frame egomotion are combined to form local five frame trajectories during test. $\otimes$ The sequence length is equal to three. The $k$ value denotes the percentage of hidden layer neurons used for egomotion prediction.}
\label{tab:kitti-ego}
\end{table}

The existing methods vary on the formation of the local five frame trajectory from the predicted egomotion during test. This is outlined in the caption of Table~\ref{tab:kitti-ego}. We present our results when the local five frame trajectories are formed by i) egomotion predictions from the central frame of the five frame snippet~\cite{zhou2017unsupervised} and ii) combination of frame-by-frame egomotion predictions~\cite{godard2018digging}. For the latter, the proposed model performs similar to the state of the art~\cite{yin2018geonet} and achieves the lowest ATE for Sequence 09 of the KITTI odometry split test set. When testing for robustness of the learned sparse representation, the proposed method achieves state-of-the-art egomotion prediction on Sequence 09 using only 5\% of the hidden layer neurons. It should be noted that despite using only local five frame snippets as input, the proposed model outperforms (Sequence 09) the ORB-SLAM (full) method that uses the complete sequence for pose estimation using global optimization steps.



\subsubsection*{Virtual KITTI}
The VKITTI egomotion split has 1679 training frames and 447 test frames. We use the relative pose error (RPE)~\cite{sturm2012benchmark} metric for evaluation of egomotion prediction, in accordance with the state of the art on the same split~\cite{tung2017adversarial}.

\begin{table}[h]
\centering
\begin{tabular}{lcc}
\hline
{Method} & {Trans. error} & {Rot. error} \\
\hline
Jaegle et al.~\cite{jaegle2016fast} & 0.4588 & 0.0014 \\
Tung et al~\cite{tung2017adversarial} & 0.1294 & 0.0014 \\
Lee and Fowlkes~\cite{lee2018cemnet} & \textbf{0.0878} & 0.0781 \\
Ours & 0.1769 & \textbf{0.0011} \\
Ours (Sparse $k=25\%$) & 0.1769 & \textbf{0.0011} \\
Ours (Sparse $k=20\%$) & 0.1788 & \textbf{0.0011} \\
Ours (Sparse $k=15\%$) & 0.1900 & \textbf{0.0011} \\
Ours (Sparse $k=10\%$) & 0.2849 & 0.0018 \\
Ours (Sparse $k=5\%$) & 0.6973 & 0.0026 \\
\hline \\
\end{tabular}
\caption{RPE comparison on the VKITTI egomotion split.}
\label{tab:vkitti-ego}
\end{table}

Table~\ref{tab:vkitti-ego} compares the proposed model against the existing methods in terms of RPE on the complete sequence of VKITTI egomotion split. The proposed model outperforms the state of art in terms of rotation error. The robust continuous egomotion estimation method by Jaegle et al.~\cite{jaegle2016fast} serves as the baseline.
The relatively large translation error maybe due to the fact that small VKITTI egomotion training set does not provide enough variations of egomotion to learn sparse features that generalize well to unseen test data. This reflects in the sparsity evaluation, as at least 20\% hidden neurons are required to predict egomotion without translation performance drop, whereas the basis learnt from KITTI data requires only 5\% active neurons to maintain performance. However, rotation accuracy is still better than the existing methods using only 15\% of the hidden neurons. This implies the usefulness of sparse representation for prediction accuracy. 

Lee and Fowlkes solves a similar formulation, including depth into the optimization procedure for egomotion prediction~\cite{lee2018cemnet}.
Tung et al.~\cite{tung2017adversarial} proposed an adversarial framework to probabilistically predict depth and camera pose. Notably, all of these methods use optic flow for egomotion prediction, either as input or using a separate optic flow prediction network. Additionally, the methods by Tung et al.~\cite{tung2017adversarial} and Lee and Fowlkes~\cite{lee2018cemnet} require depth and camera pose input for training. The proposed MFG network model uses ground truth camera pose for training.

\subsection{Object motion perception}
\begin{figure}[t]
\begin{center}
\includegraphics[width=1\linewidth]{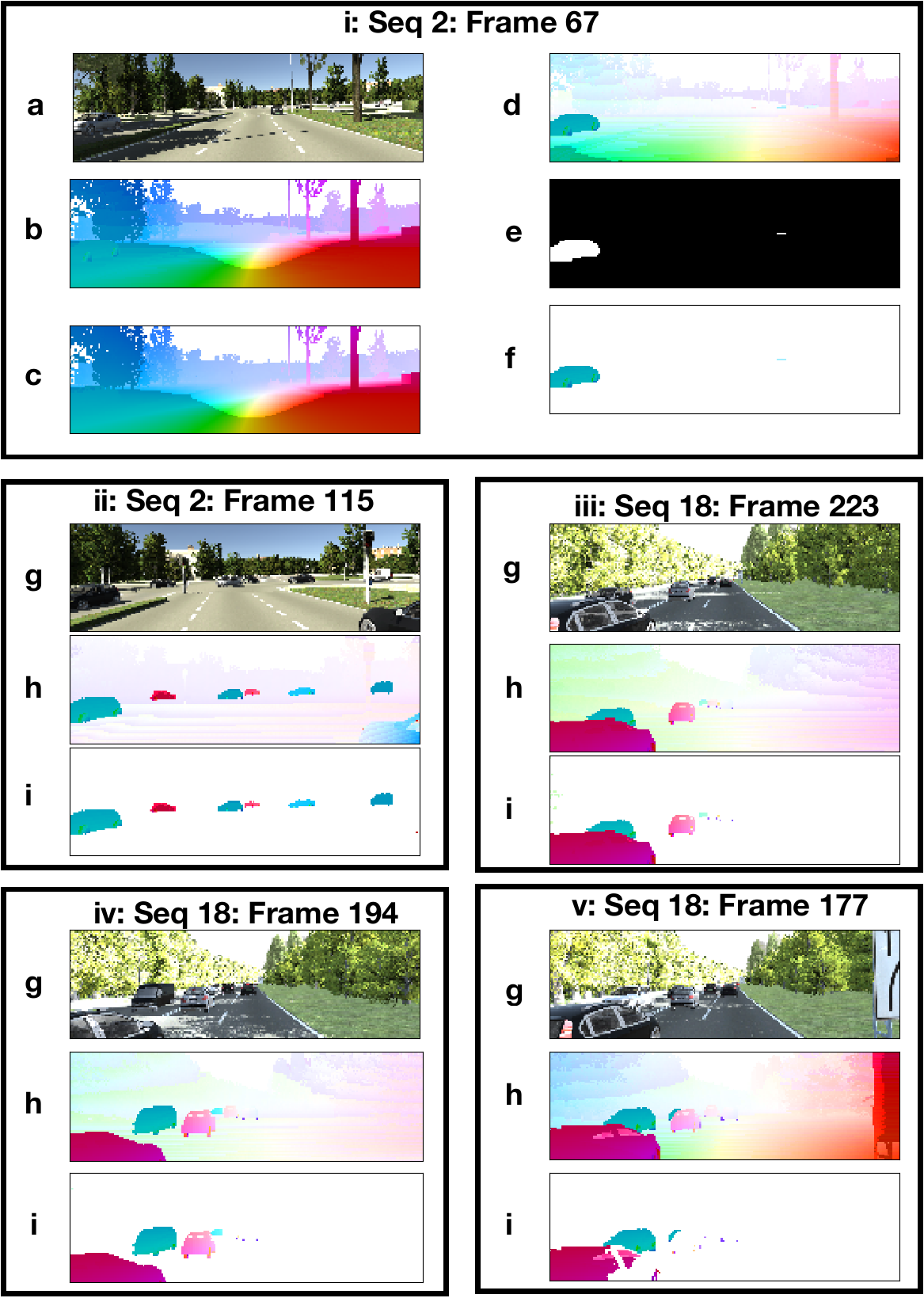}
\end{center}
\caption{Object motion extraction results on VKITTI object-motion split test set. In box i, a-f are RGB image, optic flow ($t \rightarrow t+1$), predicted motion field, predicted residue image velocity, predicted dynamic object mask, and predicted object velocity. For boxes ii-v, g is RGB image, h is predicted residue image velocity, and i is predicted object velocity.}
\label{fig:object-motion}
\end{figure}
The VKITTI object-motion split has 1554 training frames and 572 test frames. The training procedure for this task is identical to the egomotion tasks. For the object mask prediction pipeline from the residual flow, we tune $\theta_d$ parameter to 1.5 and $\theta_p$ parameter to 0.5. The value of $Threshold_D$ parameter is set to 0.2, i.e. objects closer than 20\% of the maximum depth (655 meters) will be detected. These are held fixed for all frames. 

Figure~\ref{fig:object-motion} depicts the predicted object motion masks and the extracted object motion flow vectors for five different frames of the VKITTI object-motion split. The results show that the proposed object motion extraction method is effective in finding moving objects by automatically suppressing the background residual flow vectors, even when the background residual flow is large. For instance, in frames i, iii, and v, nearby background objects (i.e. stationary) create significant residue flow. Although, the proposed object velocity extraction method is affected when the velocity differences between the moving objects are too large, which can be seen in frame v.

\subsection{Egomotion basis}

\begin{figure}[t]
\begin{center}
\includegraphics[width=1\linewidth]{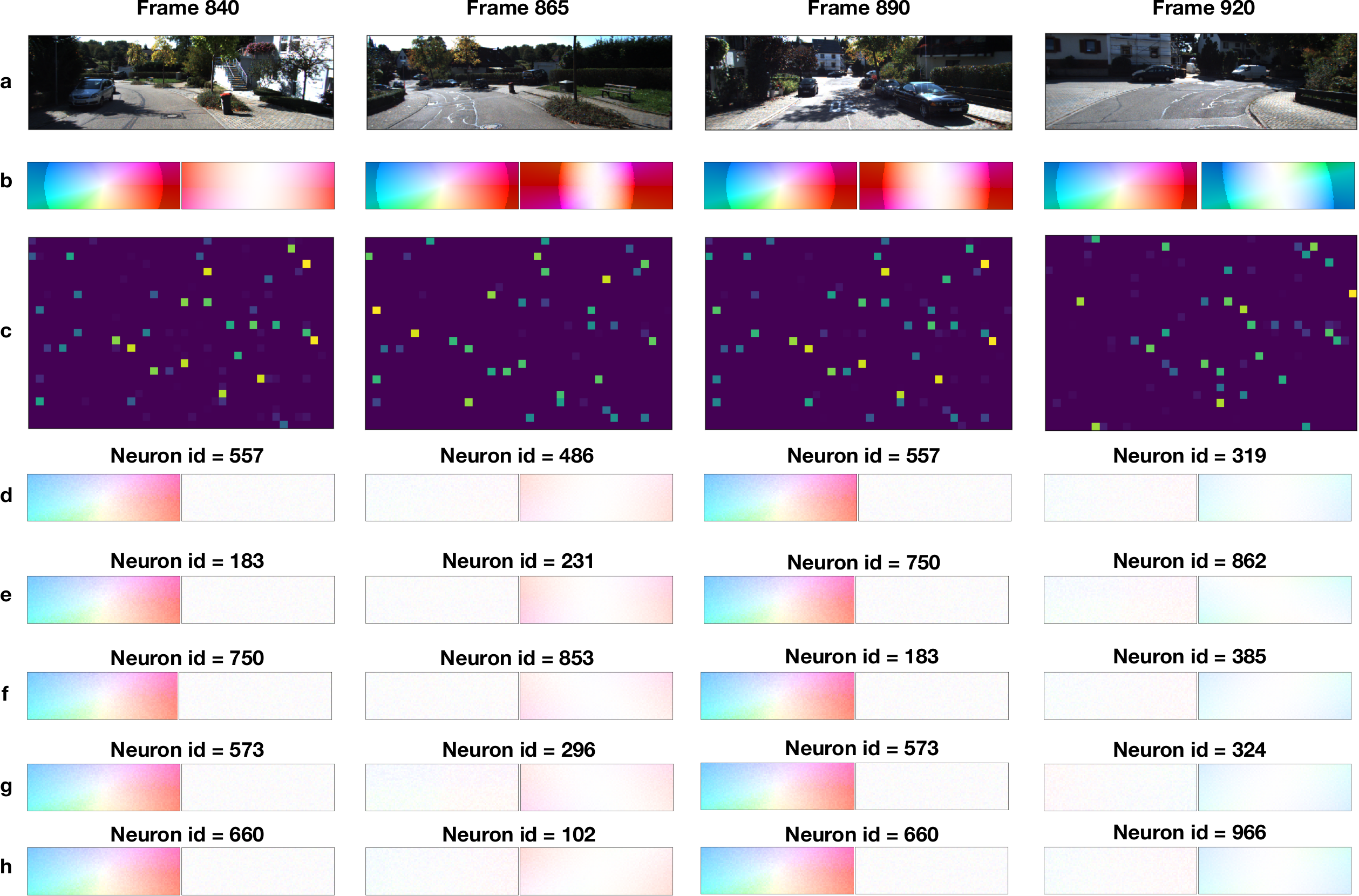}
\end{center}
\caption{The MFG network representation of egomotion $t\rightarrow t+1$ using sparsely activated basis features on KITTI odometry split test set. Each column shows the predicted motion fields and the activated components for a source frame. a) The source RGB frame, b) predicted translation-only (left, unit depth) and rotation-only (right) motion fields, c) unit activation heat map of the hidden layer neurons (brighter color means more active), d-h) Decoded translation and rotation flow fields by the five most active neurons in the hidden layer for each source frame.}
\label{fig:basis}
\end{figure}
During training, the hidden layer neurons of the MFG network learns to selectively respond to specific combinations of observer translation and rotation. Simultaneously, the decoder network learns the egomotion fields corresponding to the selectivity of individual neurons.
The decoder weights of the hidden neurons form a basis to represent egomotion.

Figure~\ref{fig:basis} depicts the representation of egomotion using a sparse population of hidden layer neurons on the KITTI odometry split test set. We select four frames 840, 865, 890,and 920 of Sequence 09, where the car is prominently going forward, left, forward, and right. Figure~\ref{fig:basis}(b) shows translation-only and rotation-only motion fields predicted by the network and Figure~\ref{fig:basis}(c) shows the activation maps of the hidden layers for each of these frames. 

For frames 840 and 890, where the car is mainly going forward, the translational component of motion is more prominent and the activation map for these two frames are almost identical. We show this further by picking out the five most active hidden layer neurons for both frames (Figure~\ref{fig:basis}(d-h)), which shows that the exact same five neurons are the most active for both frames. The decoder weights connected to these neurons represent a large forward translation along the Z-axis and little rotation representation.
The similarity of activation maps for frames 840 and 890 show the consistency of the representation learned by the MFG network for similar types of egomotion. It also implies that the learned basis vectors are not redundant.

For frame 865, where rotation of the car toward left is more prominent than frames 840 and 890, the activation map of the hidden layer neurons changes to a new state. The five most active neurons for this frame are more selective for anticlockwise rotation about the Y-axis and have little translation representation. For frame 920, where the car is slowly turning right, the activation map changes to be largely different. The five most active neurons for this source frame represent clockwise rotation about the Y-axis and have little translation representation. Although, in both frames 865 and 920, the car is still translating in the forward direction, and observedly, the most active neurons of frame 840/890 representing forward translation, map to the some subordinate activations for frames 865 and 920.

%% file: section_6.tex
In this work, we present a sparse asymmetric autoencoder with a fully convolutional encoder and a linear decoder that learns to predict egomotion with high accuracy from noisy optic flow input in dynamic scenes, while simultaneously perceiving object motion. The network is trained with constraints to learn a sparse latent space representation of egomotion with meaningful basis motion fields for camera translation and rotation. The contributions of the proposed approach regarding the existing state of the art are i) a method to directly predict camera rotation and translation from noisy optic flow marginalizing motion discontinuities from dynamic objects and variations of scene depth, ii) a sparse latent space representation of egomotion that uses only about 5\% of the hidden layer neurons for state-of-the-art egomotion prediction, and iii) dissociation of ego and object motion components across the entire image to predict fine image velocities due to independent object motion.

These contributions are useful for monocular VO estimation in dynamic scenes. Monocular VO methods that incorporate depth in the prediction loop for egomotion have scale ambiguity issues~\cite{poddar2018evolution,jaegle2016fast}. 
Our approach eliminates the need for simultaneous depth computation for egomotion prediction by using sparse representations that marginalize the depth component of input flow. It also demonstrates that marginalization of the depth component does not adversely affect the accuracy of egomotion prediction. Another advantage of our approach is that sparse representations are robust to outliers, such as independent object motion. While robust optimization procedures mask the outliers~\cite{jaegle2016fast}, our approach constructs egomotion component for dynamic regions of the scene as well, which enables computation of image velocities due to object motion.

The hidden layer neurons learn selectivity to a particular translation and rotation. It is not only a useful feature representation for machine learning, but it might be also efficient for computations as the brain uses sparse representations for many different tasks~\cite{lennie2003cost}. Previously sparse representations of egomotion components were found in the motion processing pathway of the monkey visual cortex~\cite{beyeler2017sparse}. There are two other similarities between our model and what is known about visual motion processing in the brain, i) use of vestibular (pose) information for representation of egomotion~\cite{takahashi2007multimodal} and ii) compensation of egomotion in the visual cortex to extract independent object velocities~\cite{warren2009optic}. It is important that the proposed approach shares some of the computational principles of visual motion processing in the brain and achieves state-of-the-art egomotion performance on real/realistic datasets.

\section*{Acknowledgement}
This work was supported in part by NSF grants IIS-1813785, CNS-1730158, and IIS-1253538.